\documentclass[conference, a4paper]{IEEEtran}
\IEEEoverridecommandlockouts
\usepackage{amsmath,amssymb,amsfonts}
\usepackage{algorithmic}
\usepackage{graphicx}
\usepackage{textcomp}
\usepackage{xcolor}
\def\BibTeX{{\rm B\kern-.05em{\sc i\kern-.025em b}\kern-.08em
    T\kern-.1667em\lower.7ex\hbox{E}\kern-.125emX}}

\usepackage[hyphens]{url}
\usepackage{booktabs}
\usepackage{algorithm}
\usepackage{algorithmic}
\usepackage{placeins}
\usepackage{multirow}
\usepackage{cleveref}
\crefformat{footnote}{#2\footnotemark[#1]#3}
\usepackage[style=numeric,sorting=none]{biblatex}
\begin{document}

\title{GADformer: A Transparent Transformer Model for\\Group Anomaly Detection on Trajectories\\
}

\author{
\IEEEauthorblockN{Andreas Lohrer, Darpan Malik, Claudius Zelenka, Peer Kröger}
\IEEEauthorblockA{\textit{Information Systems and Data Mining}, \textit{Kiel University}\\
Kiel, Germany \\
\{alo,cze,pkr\}@informatik.uni-kiel.de, stu225397@mail.uni-kiel.de}
}

\maketitle

\begin{abstract}
Group Anomaly Detection (GAD) identifies unusual pattern in groups where individual members might not be anomalous. This task is of major importance across multiple disciplines, in which also sequences like trajectories can be considered as a group. As groups become more diverse in heterogeneity and size, detecting group anomalies becomes challenging, especially without supervision. Though Recurrent Neural Networks are well established deep sequence models, their performance can decrease with increasing sequence lengths. 
Hence, this paper introduces GADformer, a BERT-based model for attention-driven GAD on trajectories in unsupervised and semi-supervised settings. We demonstrate how group anomalies can be detected by attention-based GAD. We also introduce the Block-Attention-anomaly-Score (BAS) to enhance model transparency by scoring attention patterns. In addition to that, synthetic trajectory generation allows various ablation studies. In extensive experiments we investigate our approach versus related works in their robustness for trajectory noise and novelties on synthetic data and three real world datasets.
\end{abstract}


\section{Introduction}
\label{01.intro}

Group Anomaly Detection (GAD) is an important task across many disciplines and domains like computational fluid dynamics and computer vision~\cite{10.5555/2986459.2986579,DBLP:conf/pkdd/ChalapathyTC18}, 
mobility~\cite{10.1145/3430195,10.1145/3557991.3567801},  physics~\cite{10.5555/3023638.3023684,PhysRevD.101.075042}, social networks~\cite{Yu2014} and many more. In these domains, GAD is suitable for various types of group anomalies. Since group member instances can be an arbitrary representation, the GAD paradigm also applies to the detection of anomalous sequences like trajectories, to which~\cite{Foorthuis2020OnTN} refers to as collective anomalies. Especially in the spatio-temporal domain, Trajectory Anomaly Detection is a common task to reveal abnormal behavior as the authors~\cite{Wang2020UnsupervisedLT,7966366,10.1145/3430195} confirm.

\begin{figure}[hbt!]
\label{fig:TrajectoryInputRepresentation}
\includegraphics[width=0.5\textwidth]{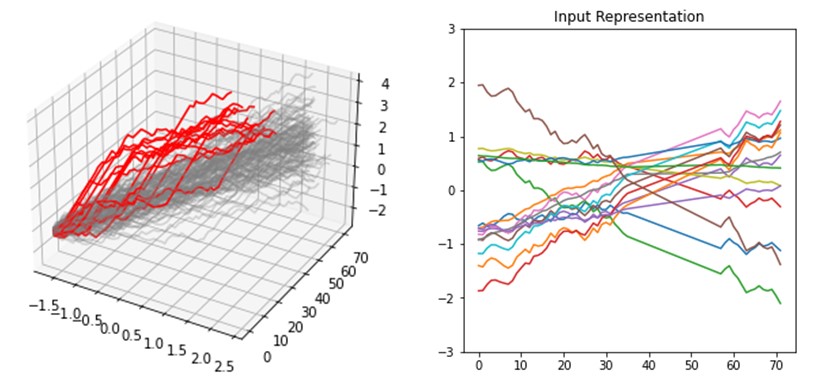}\centering
\caption[]{GADformer trajectory representations on synthetic trajectory data - on the left: 72 raw trajectory steps~$p_{n}$ (=group members~$o_{n}$) as part of gray normal or red abnormal groups (=individual trajectories~$\Lambda_{m}$); on the right: trajectory step embeddings~$e_{n}$ of one individual trajectory (=group$~\mathcal{G}$).}
\end{figure}

However, although sequential coordinates of a trajectory obviously represent a group structure, the detection of individual anomalous trajectories has not been addressed as a group anomaly detection problem yet and not at all with transparent multi-head attention.

The current state of the art approaches for anomaly detection on trajectories are recurrent neural networks (RNNs) like LSTMs~\cite{10.1162/neco.1997.9.8.1735} and GRUs~\cite{Cho2014GRU,7966366}, but the potential of deep learning methods for Group Anomaly Detection has rather been sparsely investigated. So far GAD tasks have more likely been solved by generative topic models ~\cite{Xiong2011HierarchicalPM,Yu2014,10.5555/2986459.2986579} or SVM-based methods~\cite{6790022,10.5555/3023638.3023684,10.1145/3557991.3567801}. 
Despite recent advances, deep generative models got only involved in form of Adversarial Autoencoders (AAEs) and Variational Autoencoders (VAEs)~\cite{DBLP:conf/pkdd/ChalapathyTC18} to perform GAD for images. \cite{10.1145/3430195} offers different machine learning based algorithms to detect anomalous groups of multiple trajectories, but does not identify the detection of individual anomalous trajectories (a sequence of group members) as group anomaly detection problem.

However, anomalous behavior is not ensured to just appear within short trajectory segments. It is challenging for recurrent neural networks, sometimes even LSTM and GRU, to learn very long-term dependencies~\cite{7966366}.

A further challenge for deep learning based group anomaly detection on trajectories is, that although trajectory data is highly available, it is rather weakly labeled or does not overcome the nonground-truth problem~\cite{Wang2020UnsupervisedLT} at all.

In order to tackle these challenges we introduce our approach GADFormer, a BERT\cite{DBLP:conf/naacl/DevlinCLT19} based architecture with transformer\cite{10.5555/3295222.3295349} encoder blocks for attention-based group anomaly detection on trajectories. Extending the idea of~\cite{10.5555/3295222.3295349} to optimize also image, audio or video sequence tasks by their transformer approach, we identify transformer based models for a sequence of trajectory points/segments as group member instances of a group anomaly detection task as similarly beneficial. Our model can be trained in an unsupervised as well as in a semi-supervised setting so that there is no or only reduced need for labeled trajectories. Furthermore, we introduce a Block Attention-anomaly Score (BAS), which allows us to provide an transparent view to the capability of the transformer encoder blocks to distinguish normal from abnormal trajectory attention matrices. We show with extensive experiments on synthetic and real world datasets that our approach is on par with the state of the art methods like GRU or MainTulGAD, an adapted version of~\cite{ijcai2022p274} for GAD.

Hence, the contributions of our work can be summarized as follows:

\begin{itemize}
\item Transformer-Encoder-architecture capable to perform attention-based group anomaly detection in an unsupervised and semi-supervised setting. 
\item Identification of the detection of individual anomalous trajectories as Group Anomaly Detection problem for BERT based transformer models. 
\item Block Attention-anomaly Score for group anomalies among aggregated attention pattern of multiple attention heads providing transparency for model inspection.
\item Extensive ablation and robustness studies addressing noise, novelties and standard deviation.
\end{itemize}

The remainder of this work is structured as follows. Section~\ref{02.definitions} introduces the a formal description of the addressed problem before in Section~\ref{03.methodology} the architecture, training and model transparency of GADFormer is proposed. The experiments in Section~\ref{04.experiments} demonstrate relevance and suitability of our approach across multiple domains and Section~\ref{05.relatedwork} distinguishes our approach from related work. A final summary of the paper as well as an outline to future work is given by Section~\ref{06.conclusion}.

\section{Preliminaries and Problem Definition}
\label{02.definitions}

This section provides preliminary terminology and definitions used in this work if not referenced otherwise.

\subsection{Preliminaries}
\noindent\textit{Group Anomaly Detection (GAD)} aims to identify groups that deviate from the regular group pattern\cite{DBLP:conf/pkdd/ChalapathyTC18}.\\
\noindent \textit{Group} is a set or sequence of at least two group member instances.\\
\noindent \textit{Group Member Instance} is an arbitrary data entity described by a n-dimensional feature vector as part of a group.\\
\noindent \textit{[Group Anomaly or] Collective Anomaly} refers to a collection of data points that belong together and, as a group, deviate from the rest of the data.\cite{Foorthuis2020OnTN}

\subsection{Problem Definition}

The definitions for GAD align with~\cite{DBLP:conf/pkdd/ChalapathyTC18} for deep generative models, but got partially a different notation to emphasize its suitability for group anomaly detection on individual trajectories. The GAD problem is described as follows:

Let $x_{n} \in X$ be an instance with $X = (x_{1}, x_{2}, x_{3}, ..., x_{N})$ and $x_{n} = (a_{1}, a_{2}, a_{3}, ..., a_{V})$ with attribute $a_{v} \in \mathcal{F}$, the feature space, with 

\begin{equation*}
a_{v} =
\left\{
\begin{array}{l}
continuous, a_{v} \in \mathbb{R} \\
discrete, a_{v} \in \mathbb{N} \\
categorical, a_{v} \in \{0, 1\} 
\end{array}
\right.
\end{equation*}

Be $x_{n_{m}}$ a group member instance $o_{i}$ of the $m$th group $\mathcal{G}_{m}$ with 

\begin{equation}
\mathcal{G}_{m} = (o_{1}, o_{2}, o_{3}, ..., o_{N_{m}})
\end{equation}

and $\mathcal{D}_{GAD}$ a group anomaly detection dataset, which is a set $\mathcal{G}$ of all groups:

\begin{equation}
\mathcal{D}_{GAD} = (\mathcal{G}_{1}, \mathcal{G}_{2}, \mathcal{G}_{3}, ..., \mathcal{G}_{M})
\end{equation}

The objective of the group anomaly detection task is to distinguish normal in-distribution groups from abnormal out-of-distribution groups $\mathcal{G}$$_{\mathcal{A}}$ with the help of a pseudo group $\mathcal{G}^{(ref)}$ as an approximated reference for normal in-distribution groups. Therefore, a characterization function $f$ with 

\begin{equation}
\label{eq:GADcharacterizationFunction}
f_{\Theta}: \mathbb{R}^{N_{m}\times V} \rightarrow \mathbb{R}^{D}
\end{equation}

and an aggregation function $g$ with

\begin{equation}
\label{eq:GADaggregationFunction}
g_{\phi}: \mathbb{R}^{D} \rightarrow \mathbb{R}^{D}
\end{equation}

compose to 

\begin{equation}
\label{Eq:Gref}
\mathcal{G}^{(ref)} = g_{\phi}(f_{\Theta}(\mathcal{G}))
\end{equation}

where $f_{\Theta}$ maps the groups $\mathcal{G}_{m}$ to $D$-dimensional feature vectors representing the relationship characteristics of its group members $o_{i}$ and $g_{\phi}$ aggregates them to one $D$-dimensional feature vector representing one reference $\mathcal{G}^{(ref)}$ for the distribution of normal groups.

Finally, the abnormality of a group is defined by a group anomaly score $y_{score}$ measuring the deviation by a distance measure $d(\cdot,\cdot) \geq 0$, between $\mathcal{G}_{m}$ and the normal group reference $\mathcal{G}^{(ref)}$. Thus, the abnormality score $y_{score}$ is defined as follows:

\begin{equation}
\label{Eq:yscore}
y_{score} =  d(\mathcal{G}^{(ref)}, \mathcal{G}_{m})
\end{equation}

whereby the decision between normal and abnormal groups is defined by a threshold $\gamma$ with 

\begin{equation}
y_{label} =
\left\{
\begin{array}{l}
    1, \quad y_{score} \geq \gamma \\
    0, \quad otherwise
\end{array}
\right.
\end{equation}

Having the group anomaly detection problem described according to ~\cite{DBLP:conf/pkdd/ChalapathyTC18}, we also elaborate on the task of trajectory anomaly detection aligning with the notations of ~\cite{10.1145/3430195} with their slightly different problem of group trajectory anomaly detection instead of the here described individual trajectory anomaly detection:

A trajectory point $p$ is defined as 
\begin{equation}
p = (a_{1}, a_{2}, a_{3}, ..., a_{V})    
\end{equation}

A trajectory point embedding $e$ is defined by input embedding $h_{1}$ (cf. Figure~\ref{fig:TrajectoryInputRepresentation}) for each point $p$ as 
\begin{equation}
\label{eq:ie}
e = h_{1}(p)
\end{equation}

Since word sentences and trajectories can both be considered as sequences we create BERT-based embeddings~\cite{DBLP:conf/naacl/DevlinCLT19} for trajectories by defining trajectory segments $s_{i}$ with e.g. $s_{1}=(e_{1},e_{2}), s_{2}=(e_{3},e_{4}), ...$ for $S$ segments of segment length $L_{s}=2$, which represent local sequences within a trajectory. In addition to that, each segment $s_{i}$ is mapped to a segment embedding~$h_{2}$, which acts as an offset for each related trajectory point embedding~$e_{n}$. The sum of both, segment member $e_{n}$ and segment embedding~$h_{2}$, is denoted as trajectory segment part $sp$ with
\begin{equation}
\label{eq:oe}
sp_{i,n} = e_{n} + h_{2}(s_{i})
\end{equation}

A trajectory $\Lambda$ is defined as 
\begin{equation}
\Lambda_{m} = (sp_{11}, sp_{12}, sp_{23}, sp_{24}, ..., sp_{SN_{m}})    
\end{equation}

A trajectory dataset $\mathcal{D}_{Traj}$ is defined as 
\begin{equation}
\mathcal{D}_{Traj} = (\Lambda_{1}, \Lambda_{2}, \Lambda_{3}, ..., \Lambda_{M})
\end{equation}

By considering the task of detecting abnormal individual trajectories as group anomaly detection problem, the following associations are identified:

A trajectory $\Lambda_{m}$ applies to the semantic of a group $\mathcal{G}_{m}$ by considering trajectory segments $s$ in form of its segment parts $sp$ as group members $o$. They are represented by point embeddings $e$ adding a shared segment embedding $h_{2}$ as offset~(cf. Eq.~\ref{eq:oe}). The embeddings~$e$ represent trajectory points~$p$ and a trajectory point~$p_{i}$ is associated with an instance~$x_{n}$.

Thus, individual abnormal trajectories can be detected similarly to the group anomaly detection problem (cf. Eq.~\ref{Eq:Gref} and Eq.~\ref{Eq:yscore}) as follows:

\begin{equation}
\Lambda^{(ref)} = g_{\phi}(f_{\Theta}(\Lambda))
\end{equation}

\begin{equation}
    \hat{y}_{score} =  d(\Lambda^{(ref)}, \Lambda_{m})
\end{equation}

After revealing the associations between the GAD approach of~\cite{DBLP:conf/pkdd/ChalapathyTC18} and our approach for trajectories, also our proposed GADFormer approach (cf. Section~\ref{03.methodology}) shows the potential to be trained for each arbitrary group anomaly detection problem on sequences or non-ordered sets, as far as a group to member relationship exists.

\section{GADFormer}
\label{03.methodology}

\begin{figure*}[hbt!]
\centering
\includegraphics[width=\textwidth,trim={0cm 0.5cm 0cm 0cm},clip]{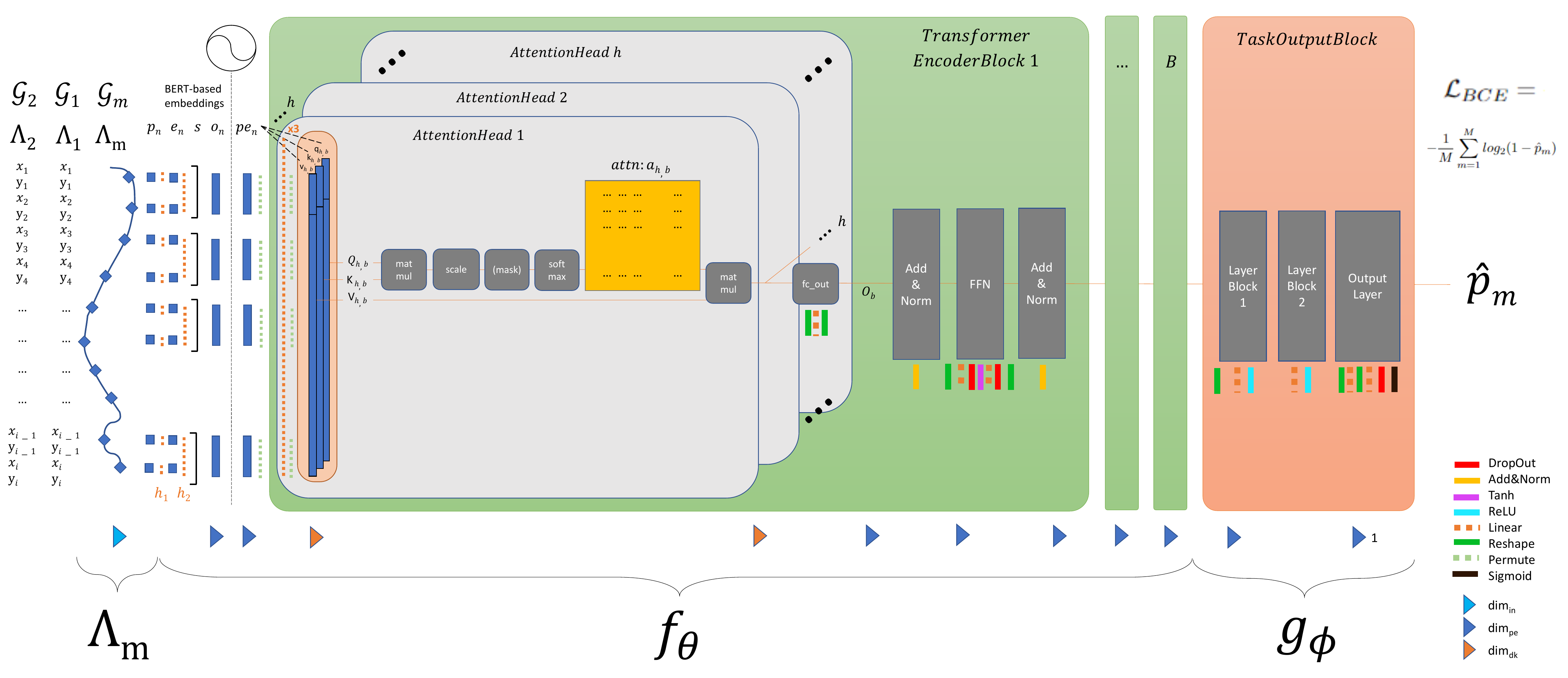}
\caption[]{GADFormer architecture overview.}
\label{fig:GADFormerArch}
\end{figure*}

In this section we propose GADFormer, a deep BERT based transformer encoder model architecture for attention-based Group Anomaly Detection (GAD). After we showed in Section~\ref{02.definitions} by the example of~\cite{DBLP:conf/pkdd/ChalapathyTC18} theoretically that the GAD problem can also be applied to trajectory coordinates, we introduce in this section with GADFormer a new deep GAD model and demonstrate its performance on trajectory datasets in Section~\ref{04.experiments}. Figure~\ref{fig:GADFormerArch} provides an overview to its model architecture in combination with examples of 2D trajectory point inputs, but also high-dimensional group members (trajectory points) are possible.

\subsection{Architecture and Loss Objective}

Differently to the GAD characterization and aggregation function (cf. Eq.~\ref{eq:GADcharacterizationFunction} and Eq.~\ref{eq:GADaggregationFunction}) of the deep generative models of~\cite{DBLP:conf/pkdd/ChalapathyTC18} our deep GADFormer $\Psi$ models the characterization and aggregation functions as follows with

\begin{equation}
    \Psi: g_{\Phi}(f_{\theta}(\Lambda_{m})) \rightarrow \hat{p}_{m}.
\end{equation}

The characterization function $f_{\theta}$ of GADFormer maps the bidirectional relationships between group members $o_{i}$ of a group $\mathcal{G}$$_{m}$ (representations of segment parts $sp$ of a trajectory~$\Lambda_{m}$) to a multi-head self-attention-weight feature map $b_{\mathcal{G}_{m}}$, representing the behavior of an individual group (an individual trajectory path pattern). This is realized by a BERT~\cite{DBLP:conf/naacl/DevlinCLT19} encoder, a composition of layers ($h_{1}$ and $h_{2}$) for input embedding, positional encoding and multi-head self-attention blocks (cf. Figure~\ref{fig:GADFormerArch}) using group member embeddings $pe$ as input tokens. In order to extend the possible value range for an improved feature extraction, we replace the ReLU activation function of the standard FFN of \cite{10.5555/3295222.3295349} with Tanh.

\begin{equation}
    b_{\Lambda_{m}} = f_{\theta}(\Lambda_{m})
\end{equation}

The aggregation function $g_{\Phi}$ of GADFormer approximates instead of a distribution for normal group representations $\mathcal{G}^{(ref)}$ with distance measure $d$ a probability $p_{m}$ for abnormal group behavior (abnormal trajectory path pattern). This is realized by non-linear layer blocks (2 linear projections with ReLU and a final output layer with linear compression and Sigmoid non-linearity) as part of the task output block $g_{\Phi}$, which maps the group behavior characteristics $b_{\Lambda_{m}}$ to a task specific feature map representation $z_{\Lambda_{m}}$. 

\begin{equation}
    z_{\Lambda_{m}} = g_{\Phi}(b_{\Lambda_{m}})
\end{equation}

After compression, the sigmoid function maps representation $z_{\Lambda_{m}}$ to a probability $\hat{p}_{m}$ for group abnormality, with $\hat{p}_{m} \in [0, 0.5]$ for normal groups and $\hat{p}_{m} \in ]0.5, 1]$ for abnormal groups (trajectories).

\begin{equation}
    \hat{p}_{m} = \sigma(z_{\Lambda_{m}})
\end{equation}

Because of the rare label availability for trajectories, the loss objective of GADFormer is defined for an unsupervised and semi-supervised learning setting assuming the majority of instances to be normal. Therefore, we define the binary cross entropy loss $\mathcal{L}_{BCE}$ as our loss function (cf. Eq.~\ref{eq:loss}). We consider this loss function as a suitable choice, since entropy $H(\hat{p}_{m})$ as a measure of unpredictability is $H(\hat{p}_{m}) = 1$ when the model is most uncertain about its abnormality prediction, and $H(\hat{p}_{m}) = 0$ when the model is very certain about its abnormality prediction. 

\begin{equation}
H(\hat{p}_{m}) =
\left\{
\begin{array}{l}
    1, \quad \hat{p}_{m}=0.5 \\
    0, \quad \hat{p}_{m}=0 \land \hat{p}_{m}=1 \\
    ]0,1[, \quad otherwise 
\end{array}
\right.
\end{equation}

Due to the heavily imbalanced learning setting with a large majority of normal groups (trajectories) one can neglect the minority of abnormal groups (trajectories) and set a fix auxiliary target probability~$p_{m}=0$ for certain normal-predictions~($H(\hat{p}_{m}) = 0$) for the majority of normal group probabilities~$\hat{p}_{m}$. In case the model faces true abnormal groups, then it is rather uncertain about its decision yielding a probability close to $\hat{p}_{m}=0.5$ resulting in a high entropy loss, whereas true normal groups, on whose pattern the model is trained, result in a low entropy loss for $\hat{p}_{m} \approx 0$.

\begin{gather}
\label{eq:loss}
\begin{aligned}
    &\mathcal{L}_{BCE} = \frac{1}{M} \sum_{m=1}^{M} p_{m} log_{b}(\hat{p}_{m}) + (1 - p_{m}) log_{b}(1 - \hat{p}_{m}) \\
    & \overset{p_{m}=0}{\Longleftrightarrow} \\
    &\mathcal{L}_{BCE} = \frac{1}{M} \sum_{m=1}^{M} log_{b}(1 - \hat{p}_{m})
\end{aligned}
\end{gather}

Since in our setting the model effectively predicts only abnormality probabilities for the range of normal groups with $\hat{p}_{m}~\in~[0, 0.5]$, where $\hat{p}_{m}=0$ means that a group (trajectory) is not abnormal at all, the abnormality of a group is defined by a group anomaly score $\hat{y}_{score} = \hat{p}_{m}$ 
for our GADFormer approach.
 
\subsection{Training}
Anomalous trajectories are rare by definition and labeling by domain experts tends to be rather expensive. We address this challenge by two different learning settings which are: 1) Unsupervised learning, which requires no labels at all under the assumption that the ratio of anomalous trajectories is low and has no remarkable influence during model training. 2) Semi-Supervised Learning, which relies on verified normal samples only. As proposed in the section before, these learning settings allow us to set a fix auxiliary target probability $p_{m}=0$, so that no ground truth for abnormal trajectories is needed for the GADFormer training.
In order to let $f_{\theta}$ learn expressive representations for $g_{\Phi}$ we start the training with frozen task layers, which get unfrozen as soon as validation loss stops decreasing. Furthermore, we use early stopping, learning rate scheduling ReduceOnPlateau and RAdam for optimization. Please see our supplementary material\footnote{\label{fn:gadf}\url{https://github.com/lohrera/gadformer}} for further details. 

\subsection{Model Transparency}

Deep learning models are known to be rather complex and their training usually requires a deep understanding for its model architecture, losses, preprocessing and data distributions to take the right decisions for fine-tuning, but still then it partially remains a blackbox as more layers, blocks and parameters exist.

\begin{figure}[hbt!]
\includegraphics[width=0.48\textwidth,trim={0 0 0 0.5cm},clip]{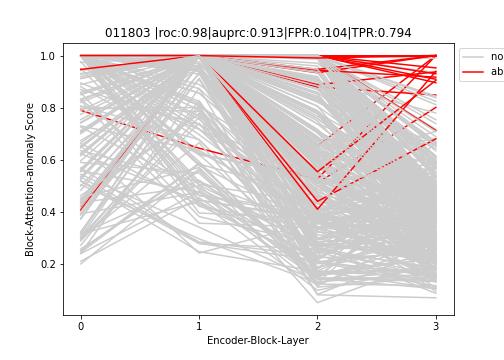}\centering
\vspace{-0.5cm}
\caption[]{BAS in case of good model performance.}
\label{fig:BAS_goodmodel}
\vspace{-0.5cm}
\end{figure}

In order to achieve a higher model transparency addressing \textit{CH6}, one of the main deep anomaly detection challenges of~\cite{10.1145/3439950}, we introduce a Block Attention-anomaly Score (BAS) for our GADFormer model. BAS can be seen as a further interpretable explanator for Model Inspection, solving a so called Open-The-Box-Problem\cite{10.1145/3236009}, which allows to indicate how each layer of the transformer encoder model contributes to distinguish inputs of different ground truth classes. Class-overlapping scores in the final layer are a potential indicator for false positives and negatives respectively. Hence, BAS enables for model inspection with the goal of identifying optimization potential in the model architecture using attention matrix scores deviating from the attention matrix score mean without plausible correlation to its ground truth and neighboring ground truths. 
%
%
%
BAS follows the assumption that in case of the aggregated attention of a group of layer heads is anomalous then also the model input, in our case the group member instances of a trajectory, is anomalous. Considering the example of Figure~\ref{fig:BAS_goodmodel} for a good model performance, this assumption holds for the majority of abnormal inputs across nearly all layers, especially for the final layer in which the amount of false positives decreases.

\vspace{-0.25cm}
\begin{figure}[hbt!]
\includegraphics[width=0.48\textwidth,trim={0 0 0 0.5cm},clip]{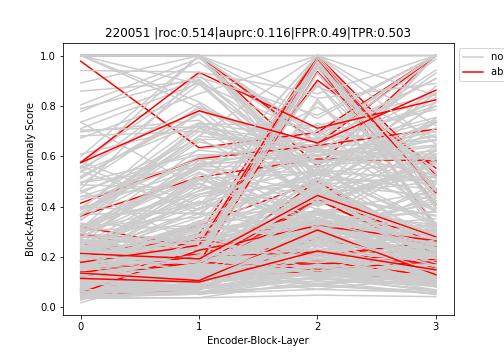}\centering
\vspace{-0.5cm}
\caption[]{BAS in case of bad model performance.}
\label{fig:BAS_badmodel}
\end{figure}

BAS represents in a transformer encoder block layer a multi-head-attention group anomaly by the ratio between distance of an aggregated block attention matrix $a_{m,b}$ and its normal block average $a_{b}$ and distance of $a_{b}$ to the average of $topN$ abnormal aggregated attention matrices $a_{topN,b}$ (cf. line 10 of Algorithm~\ref{alg:algBAS}). This allows to show the capability of a transformer encoder block layer to 1)~generally distinguish pattern of different groups~(trajectories) and 2)~to separate between normal and abnormal groups of attention head weights for individual trajectories (groups). Therewith BAS is different to the work of~\cite{Baan2019DoTA}, which aims to identify single feature-relevant attention heads by maximum attention weights and histograms instead of using average attention distances.

Since the cells of an attention matrix $a_{m,h,b}$ contain the scaled dot product of $q_{m,h}$ and $k_{m,h}$ (two projected group embeddings projected from input tokens $pe$, cf. Figure~\ref{fig:GADFormerArch}), their similarity weights the importance of the bidirectional relationship of a group member pair in different heads $h$ and with that, focuses with different views to the behavioral pattern of a group (trajectory path pattern in the context of the task of GAD on individual trajectories), whereas its concatenated projected dot product with the third projected group member embeddings $v_{m,h}$ provides the overall-importance-weighted attention output matrix $O_{m,b}$ emphasizing task relevant pattern of a concrete group $m$ (task relevant trajectory path segments of an individual trajectory $\Lambda$$_{m}$).

After emphasizing the role of the attention mechanism, we describe in Algorithm~\ref{alg:algBAS} how to calculate BAS in detail. The inputs of this algorithm are the attention matrices $a$ of the transformer encoder blocks and the euclidean distance measure. Further parameters are the ratio for the top N abnormal groups, block index~b and the amount of groups~M.

\begin{algorithm}[!htb]
    \caption{BAS Algorithm Pseudo Code}
    \label{alg:algBAS}
    \textbf{Input}: group attention matrices $a$, distance measure $d$ \\
    \textbf{Parameters}: $ratio_{topN}=0.05$, block b, groups M \\
    \textbf{Output}: block attention-anomaly scores $bas$ \\
    \vspace{-0.5cm}
    \begin{algorithmic}[1]
        \STATE $a_{m,b} = \mu(a_{m,h,b})$ 
        \IF {training}
        \STATE $tmp\_a_{b} = \mu(a_{m,b})$ 
        \STATE $topN=\lceil ratio_{topN}*M \rceil$. 
        \STATE $idx\_a_{b} = rank(d(a_{m,b}, tmp\_a_{b}), topN, dsc)$ 
        \STATE $idx\_n_{b} = idx\_all \setminus idx\_a_{b}$ 
        \STATE $global\quad a_{topN,b} = \mu(a_{m,b}[idx\_a_{b}, :, :])$ 
        \STATE $global\quad a_{b} = \mu(a_{m,b}[idx\_n_{b}, :, :])$ 
        \ENDIF
        \STATE $bas_{m,b} = min(1., \frac{d(a_{m,b}, a_{b})}{d(a_{topN,b},a_{b})})$        
        \RETURN $bas$
    \end{algorithmic}
\end{algorithm}

The average attention matrix $a_{m,b}$ for a group $m$ in heads $h$ is calculated for all attention matrices $a_{m,h,b}$ (line 1). During training also their temporary average $tmp\_a_{b}$ over all groups is calculated (line 3) to obtain their $topN$ most distant abnormal group indices $idx\_a_{b}$ (line 5), whose difference to all indices result in the remaining normal group indices $idx\_n_{b}$ (line 6). Based on these indices a global average attention heads mean for normal ($a_{b}$) and abnormal ($a_{topN,b}$) head attention averages is calculated (line 7-8) during training in order to have a solid normal and abnormal representation for distance calculation. Next, the distance between a group attention matrix $a_{m,b}$ to the normal group attention matrix average $a_{b}$ as well as the distance between the normal group attention matrix average $a_{b}$ and the abnormal group attention matrix average $a_{topN,b}$ is used to request the ratio between both which represents the Block Attention-anomaly Score $bas_{m,b}$ (line 10-11). Figure~\ref{fig:BAS_goodmodel} shows, that the first encoder block layers (0\nobreakdash-2) are not able to distinguish between normal and abnormal trajectories whereas in the last layer the amount of potential false positive scores gets less indicating a better capability of the model to attend to features of abnormal trajectories. The BAS within the layers of Figure~\ref{fig:BAS_badmodel} indicate that the model is over all layers not really able to distinguish between normal and abnormal trajectory scores, not even between the characteristics of single trajectories within the class of normal trajectories. 

In summary, the BAS provides us a view to the transformer encoder block layers for model inspection and allows us to reason reasonable changes to hyperparameters and model architecture to improve the performance of the model. Providing a further answer to "Do Transformer Attention Heads Provide Transparency in Abstractive Summarization?"\cite{Baan2019DoTA}, we are able to provide attention block transparency in terms of which degree the averaged attention of a group of self-attention-heads within one layer is normal or abnormal. 

\FloatBarrier
\section{Experiments}
\label{04.experiments}

In this section we evaluate the performance of our GADFormer approach on synthetic and real-world datasets and compare it against related works like GRU and MainTulGAD, whose approaches are state of the art methods for individual trajectory anomaly detection. For details related to datasets, architectures, hyperparameters, results and code see our supplementary material\cref{fn:gadf}. 

\subsection{Experimental Setup and Datasets}

For our experiments we tested our approach on synthetic data\cref{fn:gadf} and three real-world datasets, i.e., amazon driving routes\footnote{\label{fn:dsamazon}\url{https://github.com/amazon-science/goal-gps-ordered-activity-labels}}, Deutsche Bahn cargo container routes\footnote{\label{fn:dsdbcargo}\url{https://data.deutschebahn.com/dataset/data-sensordaten-schenker-seefrachtcontainer.html}} and brightkite checkin routes~\footnote{\label{fn:dsbrightkite}\url{https://snap.stanford.edu/data/loc-brightkite.html}} (cf. Table~\ref{tab:datasets}). The synthetically generated trajectory dataset consists of trajectory steps, one per row, where each has an id, sequence step, xcoord, ycoord, and a label whether its trajectory (=group) is anomalous (1) or normal (0).

\begin{table}
\vspace{-0.5cm}
\begin{center}
\caption{Dataset Overview.}
\label{tab:datasets}
\begin{tabular}{llrrrr}
\toprule
dataset & setting & all & normal & abnormal & trajLen \\
\midrule
synthetic\cref{fn:gadf} & unsup & 3400 & 3083 & 317 & 72 \\
synthetic\cref{fn:gadf} & semi & 3400 & 3271 & 129 & 72 \\
amazon\cref{fn:dsamazon} & unsup & 805 & 760 & 45 & 72 \\
amazon\cref{fn:dsamazon} & semi & 776 & 760 & 16 & 72 \\
dbcargo\cref{fn:dsdbcargo} & unsup & 272 & 229 & 43 & 72 \\
dbcargo\cref{fn:dsdbcargo} & semi & 245 & 229 & 16 & 72 \\
brightkite\cref{fn:dsbrightkite} & unsup & 2241 & 2033 & 208 & 500 \\
brightkite\cref{fn:dsbrightkite} & semi & 2108 & 2033 & 75 & 500 \\
\bottomrule
\end{tabular}
\vspace{-0.5cm}
\end{center}
\end{table}

All hyperparameters are empirically selected by grid search based on validation loss convergence and additionally validated by model inspection with our proposed block attention anomaly score (cf. BAS Algorithm~\ref{alg:algBAS}) in order to find ideal training parameters and a model architecture avoiding overfitting or insufficient model complexity. 

The code for GADFormer is implemented in Python utilizing PyTorch. For best possible comparison, the architecture for GRU is identical to GADFormer except using GRU-layers instead of encoder block attention layers and only input embedding (cf. Eq.~\ref{eq:ie}) instead of BERT segmentation. Our MainTul~\cite{ijcai2022p274} version (MTGAD) uses kNN-trajectory-augmentations and a student-teacher-architecture for feature extraction like the original, but adapts to sequential trajectory coordinates instead of time-dependent categorical checkins. These approaches represent the technically most related work for a fair comparison. Extended results with less related traditional methods can be found in supplementary material\cref{fn:gadf}.

\subsection{Evaluation}

For the evaluation of our model, we follow the goals of having a low miss rate (false negatives) as well as achieve as less false alerts (false positives) as possible. In addition to that, the quality of the model scores needs to be evaluated. Therefore and to be comparable to related approaches, we evaluate the model performance by AUROC and AUPRC.
\begin{itemize}
    \item $TPR = \frac{TP}{P}; FPR = \frac{FP}{N}; FNR = 1 - TPR$
    \item Precision = $\frac{TP}{TP+FP}$; Recall = $\frac{TP}{TP+FN}$
    \item AUPRC = AP = Precision vs. Recall
    \item AUROC = TPR vs. FPR
\end{itemize}

\FloatBarrier
\subsection{Results and Discussion}

\begin{figure}[hbt!]
\centering
\includegraphics[width=0.49\textwidth]{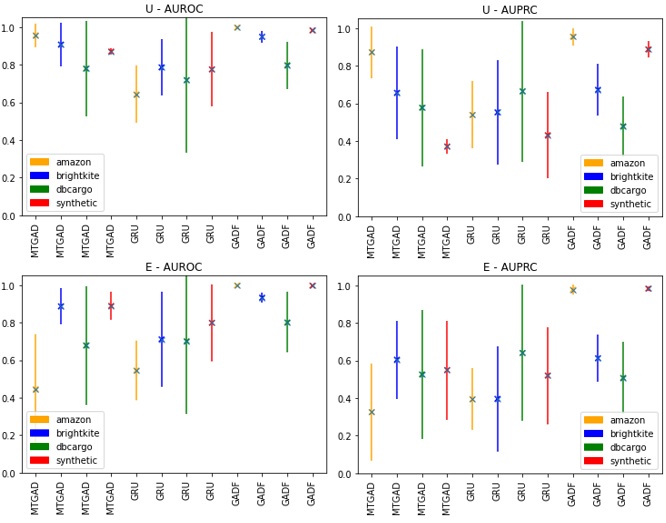}
\caption[]{Robustness results for synthetic and real world datasets of Table~\ref{tab:resultsOrigUE}. Span shows the stddev of 10 seeds; cross is the used mean performance.}
\label{fig:GADFormerErrorplots}
\end{figure}

\begin{table}
\tabcolsep=0.11cm
\begin{center}
\caption{Results on synthetic and real world datasets for (U)nsupervised- and s(E)mi-supervised setting (cf. Figure~\ref{fig:GADFormerErrorplots}).}
\label{tab:resultsOrigUE}    
\begin{tabular}{llllllllll}
\toprule
 & dataset & \multicolumn{2}{r}{amazon} & \multicolumn{2}{r}{brightkite} & \multicolumn{2}{r}{dbcargo} & \multicolumn{2}{r}{synthetic} \\
 &  & auroc & auprc & auroc & auprc & auroc & auprc & auroc & auprc \\
\midrule
\multirow[t]{3}{*}{U}  
 & GRU & 0.642 & 0.539 & 0.786 & 0.552 & 0.718 & \textbf{0.664} & 0.775 & \textit{0.431} \\
 & MTGAD & \textit{0.956} & \textit{0.872} & \textit{0.907} & \textit{0.656} & \textit{0.779} & \textit{0.577} & \textit{0.87} & 0.371 \\
 & \textbf{GADF} & \textbf{0.997} & \textbf{0.955} & \textbf{0.948} & \textbf{0.672} & \textbf{0.797} & 0.478 & \textbf{0.982} & \textbf{0.887} \\
\cline{1-10}
\multirow[t]{3}{*}{E} 
 & GRU & \textit{0.545} & \textit{0.394} & 0.711 & 0.396 & \textit{0.701} & \textbf{0.64} & 0.799 & 0.52 \\
 & MTGAD & 0.445 & 0.325 & \textit{0.887} & \textit{0.604} & 0.678 & \textit{0.526} & \textit{0.889} & \textit{0.549} \\
 & \textbf{GADF} & \textbf{0.998} & \textbf{0.976} & \textbf{0.933} & \textbf{0.612} & \textbf{0.801} & 0.507 & \textbf{0.997} & \textbf{0.982} \\
\cline{1-10}
\end{tabular}
\vspace{-0.5cm}
\end{center}
\end{table}

Our approach GADFormer (GADF) outperforms related works GRU and MainTulGAD (MTGAD) on all synthetic and real world datasets in terms of AUROC and AUPRC except for AUPRC on dbcargo for which an approach like GRU could achieve a better performance for both un- and semisupervised settings.
Considering Figure~\ref{fig:GADFormerErrorplots} and Table~\ref{tab:resultsOrigUE}, GADFormer demonstrates its stability across all datasets with lowest standard deviations over 10 seeds. For datasets amazon, brightkite and synthetic its AUROC standard deviation is close to zero. AUROC performances over 0.8 on real-world datasets in semisupervised settings highlight its relevance for real-world-domains, especially for amazon routes for which it achieved performances over 0.95 for all metrics. Also on brightkite (a dataset with long sequences of 500 steps) showed our transformer-based approach still the best performance, demonstrating its superiority against GRU and MTGAD which both are at least partially based on recurrent neural networks.
MTGAD with its self-supervised augmented kNN-trajectories and its combined student-teacher-approach of LSTM and multi-head attention shows performances close to but slightly weaker than GADFormer except for AUPRC of dbcargo. In ablation studies for detecting noise-distorted and novel anomalies shows GADFormer a comparable strong performance even for high noise and novelty ratios from 0 up to 0.5 or 0.05 respectively (cf. Table~\ref{tab:resultsNoiseUE} and Table~\ref{tab:resultsNoveltyUE}). Summarizing, compared to related work, GADFormer (GADF) can be considered as a robust approach, but despite its strong false and miss alert rates (evaluated via AUROC and AUPRC) it depends on the domains if these performances are sufficient. 


\begin{table}
\tabcolsep=0.11cm
\begin{center}
\caption{Results on synthetic dataset with noise ablations for (U)nsupervised and s(E)mi-supervised setting}
\label{tab:resultsNoiseUE}    
\begin{tabular}{llllllll}
\toprule
 & exp & \multicolumn{2}{r}{noise .0} & \multicolumn{2}{r}{noise .2} & \multicolumn{2}{r}{noise .5} \\
 &  & auroc & auprc & auroc & auprc & auroc & auprc \\
\midrule
\multirow[t]{3}{*}{U} 
 & GRU & 0.766 & \textit{0.514} & 0.731 & \textit{0.383} & 0.626 & \textit{0.165} \\
 & MTGAD & \textit{0.869} & 0.376 & \textit{0.822} & 0.256 & \textit{0.717} & 0.149 \\
 & \textbf{GADF} & \textbf{0.97} & \textbf{0.892} & \textbf{0.949} & \textbf{0.831} & \textbf{0.863} & \textbf{0.537} \\
\cline{1-8}
\multirow[t]{3}{*}{E} 
 & GRU & 0.788 & 0.585 & 0.759 & 0.479 & 0.665 & 0.223 \\
 & MTGAD & \textit{0.952} & \textit{0.766} & \textit{0.89} & \textit{0.547} & \textit{0.792} & \textit{0.316} \\
 & \textbf{GADF} & \textbf{0.989} & \textbf{0.95} & \textbf{0.98} & \textbf{0.919} & \textbf{0.944} & \textbf{0.803} \\
\cline{1-8}
\end{tabular}
\vspace{-0.5cm}
\end{center}
\end{table}

\begin{table}
\tabcolsep=0.11cm
\begin{center}
\caption{Results on synthetic dataset with novelty ablations for (U)nsupervised and s(E)mi-supervised setting.}
\label{tab:resultsNoveltyUE}   
\begin{tabular}{llllllll}
\toprule
 & exp & \multicolumn{2}{r}{novelty .0} & \multicolumn{2}{r}{novelty .01} & \multicolumn{2}{r}{novelty .05} \\
 &  & auroc & auprc & auroc & auprc & auroc & auprc \\
\midrule
\multirow[t]{3}{*}{U} 
 & GRU & 0.766 & \textit{0.514} & 0.832 & 0.585 & 0.818 & 0.496 \\
 & MTGAD & \textit{0.882} & 0.42 & \textit{0.935} & \textit{0.588} & \textit{0.923} & \textit{0.504} \\
 & \textbf{GADF} & \textbf{0.97} & \textbf{0.892} & \textbf{0.978} & \textbf{0.865} & \textbf{0.969} & \textbf{0.726} \\
\cline{1-8}
\multirow[t]{3}{*}{E} 
 & GRU & 0.788 & 0.585 & 0.849 & 0.652 & 0.841 & 0.574 \\
 & MTGAD & \textit{0.964} & \textit{0.802} & \textit{0.977} & \textit{0.867} & \textit{0.97} & \textit{0.797} \\
 & \textbf{GADF} & \textbf{0.989} & \textbf{0.95} & \textbf{0.986} & \textbf{0.921} & \textbf{0.986} & \textbf{0.841} \\
\cline{1-8}
\end{tabular}
\vspace{-0.5cm}
\end{center}
\end{table}

\section{Related Work}
\label{05.relatedwork}

Reviewing the literature for most related approaches, we could identify the following related work, which gets distinguished from our approach within this section. Instead of considering the detection of individual trajectory anomalies as a Group Anomaly Detection problem as our approach does, the vision in the works of~\cite{musleh2022let,musleh2022towards} is to observe trajectories as a NLP problem. They map trajectory coordinates to hexagon-based hexadecimal-words as input for pretrained BERT models for several tasks, but do not provide a concrete model architecture for group anomaly detection based on projected trajectory segments as BERT-based embeddings. Another work of ~\cite{ijcai2022p274} uses for the task of Trajectory-User-Linking (TUL), instead of GAD, a combination out of RNN and transformer network with cross entropy loss but compared to our approach, they do not take trajectory coordinates and segments into account and the model lacks in layer transparency. Addressing long-range trajectory anomaly detection as well the work of~\cite{9206939NFTAD} proposes an unsupervised normalizing flow (NF) model. They utilize trajectory segments and negative log-likelihood as well but use it in combination with NF-based density estimation. The work of~\cite{10.1007/978-3-030-54623-6_6} introduced the problem of group trajectory outlier detection (GTOD), which is also addressed by~\cite{10.1145/3430195}, and provide the approach CDkNN, which creates DBSCAN-based microclusters, pruned by kNN and scored with a specific pattern mining algorithm. However, both works perform anomaly detection while considering complete individual trajectories as group members, whereas we address the slightly different problem of considering single trajectory points as group members for the problem group anomaly detection. The work of~\cite{zhang2022cat} proposes a model for content-aware anomaly detection on event log messages instead of anomalous trajectories as our approach. Their approach takes additionally the content of the messages into account and allows to run it, as our approach, by the task-specific encoder-part or, differently as ours, by the typical BERT\cite{DBLP:conf/naacl/DevlinCLT19} encoder-decoder architecture. Summarizing the identified related work, there is best to our knowledge no transparent attention-based transformer-encoder-approach for group anomaly detection on coordinates-based trajectories.

\section{Conclusion}
\label{06.conclusion}





In this work we proposed GADFormer, a transformer-encoder-architecture, capable to perform attention-based group anomaly detection in an unsupervised and semi-supervised setting. We emphasized, how the detection of individual anomalous trajectories can be solved as a Group Anomaly Detection (GAD) problem for BERT based transformer models. Furthermore, we introduced BAS, a Block Attention-anomaly Score to allow model inspection for transformer encoder blocks for the task of GAD and improve with that its transparency in terms of answering to which degree the attention of the group of self-attention-heads is normal or abnormal. Extensive ablation and robustness studies addressing trajectory noise and novelties on synthetic and real world datasets demonstrated, that our approach is on par with related attention-based approaches like GRU.
Further potential for improvement could be to approximate a normal-group-distribution instead of abnormal-group-probabilities by the output-block of our model, combining the attention-based group pattern extraction of our approach and the group anomaly scoring and loss objectives of \cite{DBLP:conf/pkdd/ChalapathyTC18}.
Vice versa, with appropriate preprocessing, the performance of the GADFormer model architecture could also be evaluated on image data, audio or text data. Moreover, the reliability of the probabilities of our approach could be further investigated according to the work of~\cite{Jiang2019HowCW} as well as the relevance of single group member instances for a specific model prediction. 

\printbibliography

\end{document}